\documentclass[a4paper]{article}
\usepackage{amsmath,graphicx}
\usepackage{breakurl}
\usepackage[breaklinks]{hyperref}
\usepackage[nameinlink,capitalize]{cleveref}
\usepackage{INTERSPEECH2021}
\usepackage{cite}
\usepackage{ctable}
\usepackage[scaled=1]{beramono}
\usepackage[T1]{fontenc}

\newcommand{\mailto}[1]{\href{mailto:#1}{#1}} 

\title{End-to-End Spoken Language Understanding for Generalized Voice Assistants}
\name{Michael Saxon$^{1,2\dagger}$\thanks{$^\dagger$Work completed during author's Amazon internship.}, Samridhi Choudhary$^1$, Joseph P. McKenna$^{1\ddag}$\thanks{$^\ddag$Work completed at Amazon, currently at Google Cloud AI.}, Athanasios Mouchtaris$^1$}
\address{$^1$Alexa AI, Amazon, USA \\$^2$University of California, Santa Barbara, USA}

\email{\mailto{saxon@ucsb.edu}, \mailto{samridhc@amazon.com}, \mailto{mouchta@amazon.com}}
\newcommand{\citet}{\cite } 
\renewcommand{\paragraph}[1]{\noindent\textbf{#1}\quad}

\begin{document}

\maketitle
\begin{abstract}
End-to-end (E2E) spoken language understanding (SLU) systems predict utterance semantics directly from speech using a single model. Previous work in this area has focused on targeted tasks in fixed domains, where the output semantic structure is assumed a priori and the input speech is of limited complexity. In this work we present our approach to developing an E2E model for generalized SLU in commercial voice assistants (VAs). We propose a fully differentiable, transformer-based, hierarchical system that can be pretrained at both the ASR and NLU levels. This is then fine-tuned on both transcription and semantic classification losses to handle a diverse set of intent and argument combinations. This leads to an SLU system that achieves significant improvements over baselines on a complex internal generalized VA dataset with a 43\% improvement in accuracy, while still  meeting the 99\% accuracy benchmark on the popular Fluent Speech Commands dataset. We further evaluate our model on a hard test set, exclusively containing slot arguments unseen in training, and demonstrate a nearly 20\% improvement, showing the efficacy of our approach in truly demanding VA scenarios.
\end{abstract}
\noindent\textbf{Index Terms}: End-to-end, spoken language understanding, voice assistants, BERT, transformers, pretraining

\section{Introduction}

Spoken language understanding (SLU) systems produce interpretations of user utterances to enable interactive functions \cite{tur2011spoken}. SLU is typically posed as a recognition task, where an utterance's semantic \textit{interpretation} is populated with results from various sub-tasks, including utterance-level label identification tasks like domain and intent classification as well as sequence tagging tasks such as named entity recognition (NER) or slot filling. The conventional approach to SLU breaks the task into two discrete problems, each solved by a separately-trained module. First, an automatic speech recognition (ASR) module transcribes the utterance to text. This is then passed on to a natural language understanding (NLU) module that infers the utterance interpretation by predicting the domain, intent and slot values. Deep learning advances in both ASR \cite{hinton2012deep, graves2013speech, bahdanau2016end} and NLU \cite{xu2014contextual, ravuri2015recurrent, sarikaya2014application} have improved the performance of SLU systems, driving the commercial success of voice assistants (VAs) like Alexa and Google Home.  However, a drawback of this modular design is that the components are trained independently, with separate objectives. 
Errors encountered in either model do not inform the other;
in practice this means incorrect ASR transcriptions might be ``correctly'' interpreted by the NLU, thereby failing to provide the user's desired response.
While work is ongoing in detecting \cite{tam2014asr},  quantifying \cite{voleti2019investigating, moore2019say}, and rectifying \cite{raghuvanshi2019entity, wangasr} these ASR driven NLU misclassifications, end-to-end (E2E) approaches are a promising way to address this issue. 

Rather than containing discrete ASR and NLU modules, E2E SLU models are trained to infer the utterance semantics directly from the spoken signal \cite{haghani2018audio,lugosch2020using, palogiannidi2020end, radfar2020end, tian2020improving, chung2020semi, kim2020two, kim2020st}.  These models are trained to maximize the SLU prediction accuracy where the predicted semantic targets vary from solely the intent~\cite{lugosch2019speech, chen2018spoken}, to a full interpretation with domain, intents, and slots~\cite{haghani2018audio}. The majority of recent work on English SLU has targeted benchmark datasets such as ATIS \cite{price1990evaluation}, Snips \cite{coucke2018snips}, DSTC4 \cite{yang2017end} and Fluent Speech Commands (FSC) \cite{lugosch2019speech}, with FSC in particular gaining recent popularity. A similar collection of French spoken NER and slot filling datasets has been investigated \cite{tomashenko2019recent}. Over the last year the state-of-the-art on FSC has progressed to over 99\% test set accuracy for several E2E approaches \cite{lugosch2020using, palogiannidi2020end, radfar2020end, tian2020improving, chung2020semi, kim2020two, kim2020st}. 
However, there remains a gap between the E2E SLU capabilities demonstrated thus far and the requirements of a \textit{generalized} VA~\cite{mckenna2020semantic}. In particular, existing benchmarks focus on tasks with limited \textbf{semantic complexity} and output \textbf{structural diversity}.

Different SLU use-cases have significantly different dataset requirements and feasible model architectures. For example, controlling a set of smart appliances may only require device names and limited commands like ``on'' and ``off.'' Similarly, a flight reservation system can assume the user intends to book a flight \cite{Kuo2020}. In these settings, a restricted vocabulary and output structure is appropriate to ensure high performance. 
However, when interacting with generalized VAs like Alexa, users expect a system capable of understanding an unrestricted vocabulary, able to handle any song title or contact name. This leads to tasks with a long tail of rare utterances containing unique $n$-grams and specific slot values unseen during training, that are more \textit{semantically complex} than the tasks tackled in aforementioned benchmark SLU datasets. 
Differences in semantic complexity across datasets can be assessed using $n$-gram entropy and utterance embedding MST complexity measures \cite{mckenna2020semantic}.
Furthermore, in generalized VA tasks the output label space is countably infinite, as any arbitrary sequence of words could be a valid slot output. Thus an assumption of a simple output structure is no longer valid, making the problem \textit{structurally diverse}. 

Designing an E2E system for semantically complex and structurally diverse SLU use-cases is the focus of this work. We present a transformer-based E2E SLU architecture using a multi-stage topology \cite{haghani2018audio} and demonstrate its effectiveness in handling structurally diverse outputs, while achieving the $99\%$ accuracy benchmark for FSC. 
We use an de-identified, representative slice of real-world, commercial VA traffic to test if our model is capable of handling \textit{complex} datasets. Furthermore, we demonstrate how to leverage large-scale pretrained language models (BERT) and acoustic pretraining for increased robustness. We perform a supplementary analysis across multiple choices of differentiable interfaces for our multistage E2E setup. Finally, we show the performance of our proposed model on ``hard'' data partitions which exclusively contain slot arguments that are absent from the training data, demonstrating more robust performance in demanding general VA settings.



\begin{figure}[t!]
	\centering
	\includegraphics[width=0.72\linewidth]{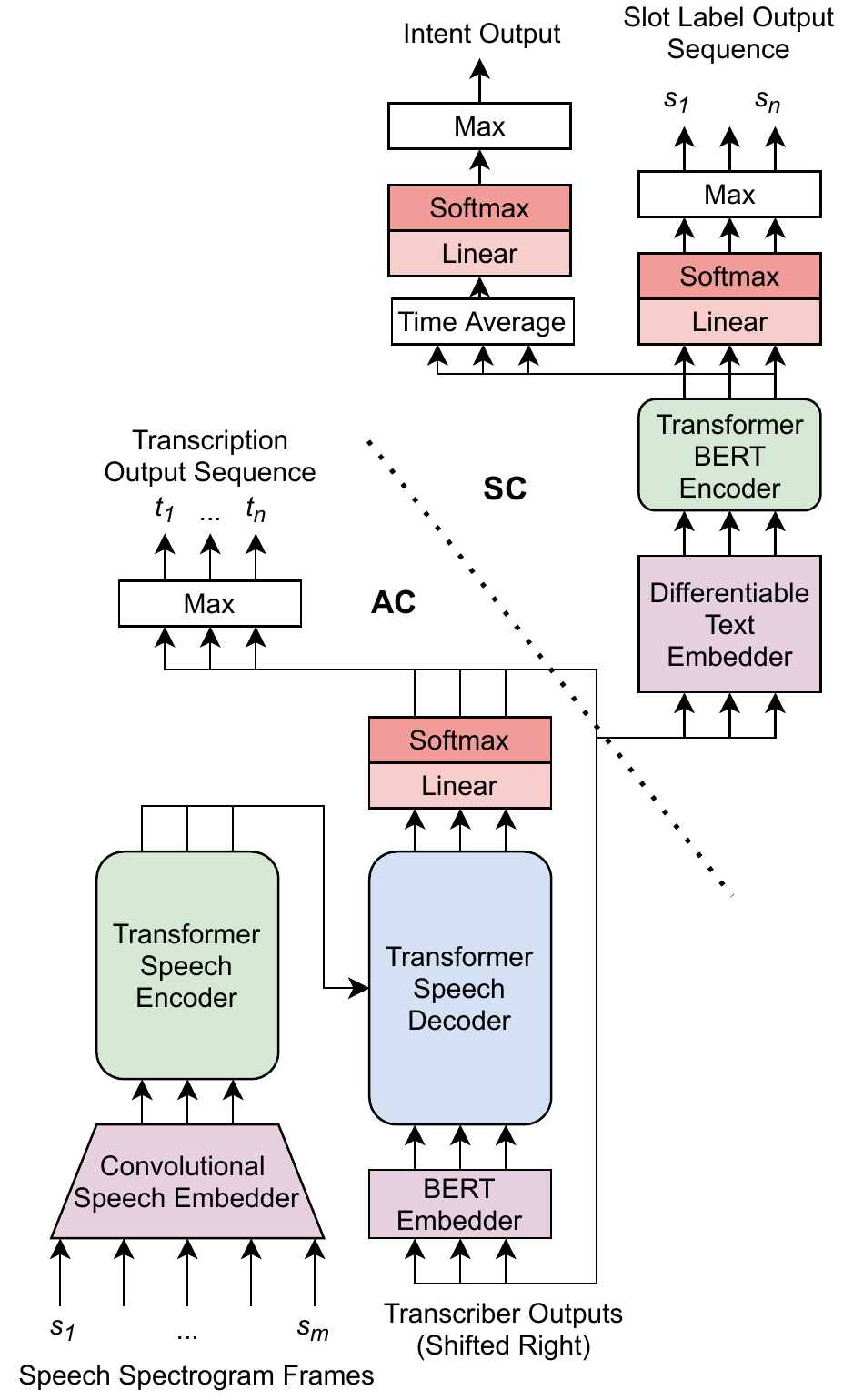}
	\caption{A diagram depicting the full E2E SLU model and the soft  Acoustic (AC) and Semantic (SC) component boundary.}
	\label{fig:model}
	\vspace{-12pt}
\end{figure}

\section{Model Architecture}
\label{sec:model}

We adopted the multistage E2E topology from \cite{haghani2018audio}, that resembles an end-to-end trainable variation of the traditional modularized SLU architecture. Due to this resemblance, we find it helpful to think of our model, shown in Figure \ref{fig:model}, as being composed of two components: an ``acoustic component'' (AC) and a ``semantic component'' (SC). The AC takes in speech spectrograms and outputs a sequence of wordpiece tokens. The SC ingests the AC's output posterior sequence and produces an utterance-level intent class and a sequence of wordpiece-level slot labels. These two components are connected by a modified embedder that is differentiable by operating on wordpiece posteriors. Thus gradients flow from SC to AC, enabling end-to-end training for the entire setup on a single training objective. The differentiable interface idea is similar to \cite{Rao_2020} except we employ it to build the SC around a pretrained neural language model. 

This architecture gives us the flexibility of still being able to produce a transcription, from which the slot values can be extracted via a slot tagger. However, unlike the modular SLU, we propagate gradients from semantic loss all the way to the acoustic input layer. Moreover, we can selectively pretrain components with different datasets and various objectives across the speech and text modalities. For example, the AC can be pretrained using non-SLU, speech-only datasets, that are often available in large quantities. Similarly, since the SC operates on wordpiece-level data, it can be designed to use a pretrained language model, in this case BERT~\cite{devlin2018bert} as a text encoder, where we attach task-specific heads to create an appropriate SC. Therefore, we are able to incorporate appropriate inductive biases in the model, by capturing both the acoustic (via AC pretraining) and linguistic information (via SC pretraining) that is difficult to learn from relatively small E2E SLU datasets.

\noindent \textbf{Acoustic Component (AC)} ---
The AC is made up of a convolutional neural network (CNN)-based time-reducing embedder, a transformer encoder, and a transformer decoder.
The input to the embedder consists of $256d$ log spectrograms with a 20 ms frame length and 10 ms frame spacing. These  frames are embedded using three $1d$ convolutional layers, with an output size 240, kernel size 4, stride 2, and ReLU activations. After embedding, a sequence of encodings corresponding to 240 ms of input audio, with a 120 ms spacing are produced. This architecture is inspired from the time-reducing convolutional speech encoders employed in \cite{radfar2020end}.
A sequence of wordpieces is then autoregressively transcribed from the encodings using a 12-layer, 12-head transformer encoder-decoder with hidden size 240, trained with teacher forcing during both pretraining and fine-tuning \cite{chiu2018state}.

\noindent \textbf{Semantic Component (SC)} ---
The semantic component is made up of four parts---a differentiable embedder, a pretrained BERT encoder, an utterance-level dense intent decoder, and a wordpiece-level dense slot sequence decoder.
The differentiable embedder performs the same function as the typical BERT embedder lookup table, but can take in uncertain posterior inputs from the AC during training, enabling end-to-end gradient flow. 
The pretrained BERT encoder is a standard 12 layer $768d$ transformer encoder, that takes in the sequence of embeddings from the differentiable embedder and outputs a sequence of encodings of equal length.
The intent decoder is a single linear layer of size $768\times N_{IC}$ (num. intent classes) that takes the time-averaged encoded sequence to generate a single intent class estimate. The slot label sequence decoder is a single linear layer of size $3072\times N_{SL}$ (num. slot labels). The input to this decoder is formulated by concatenating the top 4 BERT encoder layer outputs at each step \cite{devlin2018bert}, while the output is a slot label estimate. The final sequence of (slot label, slot value) pairs is constructed by concatenating subsequent wordpiece tokens tagged with a slot label other than null.

\noindent \textbf{Differentiable Embedders} ---
\label{sec:diffemb}
In a non E2E system, an argmax over the vocabulary length dimension could be performed on the AC output, after which the BERT lookup table would embed the transcribed word-pieces. However, this approach interrupts gradient flow, thereby rendering E2E training impossible. We experimented with three different approaches to generate differentiable BERT input encodings from the AC output posteriors. As some approaches to doing this require producing a very large internal posterior or producing large matrix multiplications (vocab size x vocab size), we analyze their impacts on both accuracy and inference speed.

\textbf{TopK:} In this approach, the posterior sequence of the embedder is sorted along the vocabulary dimension to produce a sequence of tokens of decreasing likelihood. This is followed by generating a mixture of the top-$k$ token embeddings using the embedding lookup table and the softmax values of the top-k tokens. We used $k=20$.

\textbf{MatMul:} Here, we store a $\textit{vocab size}\times\textit{embedding size}$ matrix containing the input embedding for every token in the vocabulary. With this we can easily generate a confidence-weighted mixture of all possible embeddings by multiplying this matrix by the output softmax of the embedder.

\textbf{Gumbel:} Instead of taking an argmax over the vocabulary, we instead use the Gumbel-softmax trick \cite{gumbel} to select a single word whose embedding is then passed on to the SC at each step. Gumbel-softmax helps approximate a smooth distribution for back propagation, allowing gradient flow.

\section{Methodology}
We follow a two step training approach: (1) pretrain the AC and SC layers on appropriate datasets and optimization objectives to help encode acoustic and linguistic semantic information, then (2) fine-tune the entire model end-to-end on a task-specific VA dataset. Details for our training and evaluation methodologies, datasets, and baselines are provided below.

\subsection{Pretraining}
In the pretraining stage, the AC is trained for the ASR transcription task on  460 hours of clean LibriSpeech data \cite{panayotov2015librispeech}. Rather than using typical ASR-style subwords or full words as targets, the transcriptions are converted into BERT-style wordpiece sequences using the HuggingFace {\scriptsize\texttt{bert-base-uncased}} tokenizer \cite{wolf2020transformers}. This helps us prime the AC layers to return tokens in the format that is expected at input to the BERT encoder in the SC. We use the Adam optimizer to minimize the sequential ASR cross entropy loss $ \mathcal{L}_{\textit{ASR}}$.

We built the SC around the pretrained {\scriptsize\texttt{bert-base-uncased}} model distributed by HuggingFace \cite{wolf2020transformers}. We perform no task-specific text-level pretraining beyond the cloze (masked LM) task and next sentence prediction learning that is inherent to using a pretrained BERT module \cite{devlin2018bert}. The final output linear layers (intent and slot decoders) are randomly initialized at the beginning of the end-to-end training phase.

\subsection{End-to-end training}
After pretraining, the AC and SC layers are composed such that the AC output posteriors are fed directly into the input of the SC, with the differentiable embedder acting as the embedding lookup component. This setup is trained on a three-term sum of categorical cross entropy loss for the ASR output sequence $ \mathcal{L}_{\textit{ASR}}$, the slot labels $ \mathcal{L}_{\textit{Slot}}$, and a single utterance-level intent $ \mathcal{L}_{\textit{Intent}}$.  $ \mathcal{L}_{\textit{Slot}}$ is a sequence-level target where each token in the ASR output sequence is assigned either a null output or a slot label. This three-term loss (\cref{eqn:le2e}) is minimized using Adam.

\begin{equation}\label{eqn:le2e}
	\small
	\mathcal{L}_{\textit{E2E}}=\mathcal{L}_{\textit{Intent}} +\mathcal{L}_{\textit{Slot}} + \mathcal{L}_{\textit{ASR}}
\end{equation}

\subsection{Model evaluation}
We use greedy ASR decoding to produce the output sequence of wordpieces from the AC. The inputs to SC are the output posteriors rather than the discrete word choices themselves. We perform a grid search over learning rates $\in[10^{-5},0.01]$, dropout $\in(0,1]$, and hidden layer sizes $\in\{120, 240, 400, 512\}$, as well as experiment with slanted triangular learning rate schedules and hierarchical unfreezing strategies as described in \cite{howard2018universal}, to get the best performing model. All models were trained and evaluated on EC2 instances with Tesla V100 GPUs.  In order to analyze the final SLU performance, we use three metrics:
\begin{enumerate}
	\item \textbf{Intent Classification Error Rate (ICER) }- Ratio of the number of incorrect intent predictions to the total number of utterances.
	\item \textbf{Slot Error Rate (SER)} - Ratio of incorrect slot predictions to the total number of labeled slots in the dataset.
	\item \textbf{Interpretation Error Rate (IRER)} -  Ratio of the number of incorrect interpretations to the total number of utterances. An incorrect interpretation is the one where either the intent or the slots are wrong. This ``exact match'' error rate is the strictest of our evaluation metrics.
\end{enumerate}

\subsection{Data}
We use two E2E SLU datasets for our experiments - (1) the publicly available Fluent Speech Commands (FSC) and (2) an internal SLU dataset. Additionally, we create a ``hard test set'' to assess model performance in the most demanding scenarios in generalized VA. We use the average $n$-gram entropy and Minimum Spanning Tree (MST) complexity score as described in \cite{mckenna2020semantic} to quantify their levels of semantic complexity.

\noindent \textbf{Fluent Speech Commands} ---
FSC \cite{lugosch2019speech} is an SLU dataset containing 30,043 utterances with a vocabulary of 124 words and 248 unique utterances over 31 intents in home appliance and smart speaker control. The SLU task on this dataset is just the intent classification task. It has an average $n$-gram entropy of 6.9 bits and an average MST complexity score of 0.2 \cite{mckenna2020semantic}. 

\noindent \textbf{Internal SLU Dataset} --- 
In order to analyze the effectiveness of our proposed architecture on a generalized voice assistant (VA) setting, we collect a random, de-identified slice of internal data from a commercial VA system. The data is processed so that users are not identifiable. The resulting dataset contains about 150 hours of audio, with over 100 different slot labels, dozens of intent classes and no vocabulary restrictions.  It has an average entropy of 11.6 bits and an average MST complexity of 0.52 \cite{mckenna2020semantic}. Both complexity metrics, alongside the less structurally constrained output label space, demonstrate that this task is more complex than FSC.

\noindent \textbf{Hard Subset of Internal Traffic Data} ---
In generalized VA, accuracy on semantic outliers is desirable. To assess this dynamic we produce a \textbf{hard test set} of 18k utterances from our internal  dataset. This is done by selecting utterances that exclusively contain at least one \textit{minimum-frequency bigram}, a pair of subsequent words that is not present in our training or validation sets. This test set helps us simulate how a system will perform on unforseen utterances that tend to arise in production VA.

\begin{table*}[ht!]
	\small
	\centering
		\caption{Results from Internal Traffic Dataset, for both the regular and hard test sets. Relative improvement in absolute Intent Classification Error Rate (ICER), Slot Error Rate (SER) and Interpretation Error Rate (IRER) are reported as positive deltas over the Multitask LSTM Baseline (lowest performance). }
	\label{tab:alexaresults}
	\begin{tabular}{cccccccc}
		\cline{3-8}
		\multicolumn{2}{c}{} & \multicolumn{3}{c}{Regular Test Set} &  \multicolumn{3}{c}{Hard Test Set} \\
		\hline 
		\textit{Model} & \textit{Num. Params.} & \textit{ICER}  & \textit{SER}  & \textit{IRER} &  \textit{ICER}  & \textit{SER} &  \textit{IRER}  \\ \specialrule{.1em}{.05em}{.05em}
Multitask LSTM Baseline &  45.9M  &--- & --- & --- & --- & --- & --- \\
Multitask BiLSTM Baseline & 67.3M  &+0.5 & +0.5 & --1.1 & +1.8 & +0.2 & +0.5 \\
Multitask Transformer Baseline &  101M  &+4.0 & --2.7 & --0.1 & +4.4 & --3.5 & +0.2 \\ \hline
\textbf{Ours} (No Pretraining) &  115M  &+9.1 & +37.1 & +40.5 & +11.4 & +14.2 & +18.1 \\
\textbf{Ours} (BERT, AC Pretraining) &  115M  & \textbf{+9.3} & \textbf{+37.3} & \textbf{+42.8} & \textbf{+12.9} & \textbf{+15.0} & \textbf{+18.9} \\
		\hline
	\end{tabular}
\end{table*}

\subsection{Baselines}
We design our baselines using a multitask E2E topology, defined by Haghani et al.~\cite{haghani2018audio}.
Our ability to use proven E2E models vetted on public SLU tasks such as FSC, as baselines, is hampered by the fact that they are typically designed with non-generalized VA use-cases in mind. In particular, the hard subset classification task is impossible for the models designed according to the direct or joint topologies from \cite{haghani2018audio} to perform without significant modification. Specifically, they lack the ability to select arbitrary words from the transcription vocabulary as slot values. Most high-performing models for FSC follow the direct or joint topology \cite{radfar2020end,kim2020st, lugosch2020using}. Instead, the multitask topology \cite{haghani2018audio} provides a good contrast to our proposed multistage model; both maintain the necessary capability of identifying slots by labeling a sequence of wordpieces.

We analyze three baseline multitask models, that differ only in the sequential encoder and decoder used, in particular (1) unidirectional LSTM, (2) bidirectional LSTM, and (3) transformer. All baseline models use a CNN-based speech spectrogram embedder identical to the one presented in Section \ref{sec:model}. This is followed by the speech sequence encoder using one of the three aforementioned encoder types. Finally, these encodings are decoded with task-specific heads, that consist of a dense layer for utterance level intent classification and word-level dense layer for sequential slot decoding. The final structured output for IRER evaluation contains the slot values and slot labels along with the intent label for the entire utterance. 
Our baselines allow us to evaluate both the efficacy of a multistage setup and of using a transformer based encoder-decoder with BERT.


\section{Results}
We present internal dataset results in Table \ref{tab:alexaresults}. All metrics are reported as relative improvements in percent over the simplest baseline model (the unidirectional ``multitask LSTM baseline''). 
We also report the results for our architecture with randomly-initialized AC and SC at the start of fine-tuning (No Pretraining). In this condition the model is only trained on our internal dataset, from scratch. As we can see our pretrained model, with both LibriSpeech AC pretraining and BERT language model pretraining, achieves the best performance with a 9.3\% improvement in ICER, 37.3\% in SER and a huge 42.8\% in IRER, on the ``regular'' test set.  For this table we use the best-performing Gumbel interface (\autoref{sec:interface}).

The hard test set results are especially noteworthy. While baselines struggle to correctly identify slot values at all, our model improves the hard test set IRER by $\approx 19\%$.  Many of these slot arguments are never seen in the training data, and are only correctly classified because our model is able to successfully identify which wordpieces in the output sequence should correspond to a slot value. This gain in generalization performance is a strength of our approach and demonstrates the efficacy of this architecture for complex use-cases.

Our model achieved a 0.6\% IRER (99.4\% accuracy) on the FSC dataset. While our model is designed for a  structurally diverse and semantically complex SLU use-case, it nevertheless meets the benchmark of beating 99\% accuracy on FSC, previously demonstrated in \cite{chung2020semi, lugosch2020using, tian2020improving, kim2020st, kim2020two}, and is therefore comparable to the state-of-the-art in intent-only classification performance on FSC.

\subsection{Embedder analysis}\label{sec:interface}
We evaluate all three proposed differentiable emedders, by reporting the inference speed and accuracy for models containing each. We timed the speed of inference on a single, 32-utterance minibatch on a single GPU. We also report the ICER, IRER and the hard test set IRER (h-IRER) for each interface.

\begin{table}[h!]
	\small
	\centering
	\caption{Comparing the speed and performance of the three differentiable interfaces on the ``regular'' test set.	\label{tab:interfaces} }
	\begin{tabular}{ccccc}
		\hline
		\textit{Interface} & \textit{Speed}  & \textit{ICER} & \textit{IRER}& \textit{h-IRER}  \\ \specialrule{.1em}{.05em}{.05em}
		MatMul & --- & --- & --- & --- \\
		Top20 & +10 ms & +5.1 & +2.7 & +1.5 \\
		\textbf{Gumbel} & \textbf{--16 ms} & \textbf{+7.2} & \textbf{+4.3} & \textbf{+2.4} \\
		\hline
	\end{tabular}

\end{table}

\noindent As seen in Table \ref{tab:interfaces}, the Gumbel-softmax outperforms the other interfaces both in inference speed and error rates. The improved error rates suggest that certainty in the selection of words being passed to the SC improves performance.


\section{Discussion}
\label{sec:disc}
Our approach meets the 99\% test accuracy benchmark on FSC.  However, this benchmark task is simple, with low semantic complexity \cite{mckenna2020semantic}. 
The key benefit to our approach is its ability to perform inference on a structurally diverse set of semantically complex utterances. Our multistage model, containing a differentiable interface with ASR- and NLU-level fine-tuning on task-specific data, is able to not only beat baselines on production-like structurally diverse traffic but also generalize to a hard test set uniquely composed of previously unseen slot arguments, achieving a 19\% improvement over the very poor IRER achieved by the baselines. 

We note that pretraining both the AC and SC modules only produced modest improvements over random initialization. This might be because the quantity of data provided during the fine-tuning stage is sufficiently large for achieving a good fit, providing a good sample of relevant transcriptions and interpretations. Alternatively, the pretrained representations might be too general or in the wrong domains; the audiobook speech in LibriSpeech and the massive corpora of internet text used to train BERT span diverse topics. This pretraining data may be of limited applicability to our setting when sufficient in-domain data is available. 

For scenarios where generalized VA is necessary, but less training data is available, our proposed architecture would enable using a maximum amount of semantic pretraining for each modality of the model (speech and text).
Apart from pretraining, following a multistage approach is also one of the core reasons that our model performs so well, especially for the hard test set.
By accepting transcription loss during fine-tuning, the model is constantly corrected on recognizing the lexical content of user utterances. By forming the semantic decision from these supervised transcriptions, the SC is able to directly benefit from the improved AC accuracy in a way that multistage models (such as our baselines) and direct models \cite{haghani2018audio} can not.

\section{Conclusion}

We have demonstrated the performance of a multistage transformer-based E2E SLU model that is capable of handling the output structural diversity necessary for deployment in a generalized VA setting. We have shown that this approach significantly outperforms various multitask baselines on the hardest slot classification examples characteristic of semantically complex datasets. Furthermore, we demonstrated that these gains in functionality do not come at a cost of performance on simpler SLU benchmarks. We hope for future work further exploring E2E SLU in structurally diverse, semantically complex general VA settings, especially in low-data scenarios.

\bibliographystyle{IEEEtran}

\bibliography{acl2020}

\end{document}